# Face Recognition with Machine Learning in OpenCV – Fusion of the results with the Localization Data of an Acoustic Camera for Speaker Identification

Johannes Reschke; Armin Sehr
Department of Electrical Engineering and Information Technology
Ostbayerische Technische Hochschule Regensburg
93049 Regensburg, Germany
johannes1.reschke@st.oth-regensburg.de; armin.sehr@oth-regensburg.de

*Abstract*—This contribution gives an overview of face recognition algorithms, their implementation and practical uses. First, a training set of different persons' faces has to be collected and used to train a face recognizer. The resulting face model can be utilized to classify people in specific individuals or unknowns. After tracking the recognized face and estimating the acoustic sound source's position, both can be combined to give detailed information about possible speakers and if they are talking or not. This leads to a precise real-time description of the situation, which can be used for further applications, e.g. for multi-channel speech enhancement by adaptive beamformers.

## I. INTRODUCTION

In recent times, the interest in and research on acoustic source localization and enhancement of certain sound sources has increased dramatically due to the growing desire for hands-free interaction with various devices [18]. Combining the ability to locate sound sources and to recognize possible speakers with a camera potentially enables machines to identify speakers. This makes a human-machine-interaction a lot easier, more adaptive and reliable. Comparable systems can be used for teleconferencing, smart rooms or ambient assisted living [5, 8].

Combining a microphone array's ability to locate sound sources and the intuitive way of extracting information of a webcam's image, acoustic cameras have become quite popular for many industrial segments. They compute a color-coded sound map and thus, they visualize the sound pressure levels of a user-defined field of view. This way, acoustic cameras can locate sound sources quite accurately, which is why they are often used to identify unwanted noise sources [12, 31].

Face recognition is a machine learning technique, which ideally allows detecting and identifying all faces seen in a picture or a video frame. It can be used for criminal detection, image processing, human computer interaction, etc. [33, 34]. In the early development of face recognition systems, geometric facial features, e.g. eyes, nose and mouth, were explicitly used. Properties of these features and relations (e.g. positions, distances, angles) between them were used as descriptors for face recognition [15]. Today, holistic techniques, e.g. principal component analysis (see Eigenfaces) or linear discriminant analysis (see Fisherfaces), are used to identify individuals [3, 13].

## II. BASICS OF AN ACOUSTIC CAMERA

The implementation of an acoustic camera requires a suitable microphone array as well as beamforming algorithms to locate sound sources precisely. Given both, a color coded sound map of the measured sound pressure level can be computed and displayed as seen in Figure 3 [25, 26].

### A. A suitable microphone array

In [26] a appropriate microphone array (see Figure 1) for speaker and sound source localization has been developed and verified. It can be shown that double ring arrays with an odd number of microphones on each ring are desirable for locating speech sources [20]. In this project, the inner ring has a diameter of 0.2 m, while the outer ring is twice as large. An important part of acoustic cameras is a sound analysis and visualization software [31]. This software can for example be installed on a personal computer. The connection of microphones with any computer is achieved by using a microphone amplifier and a multi-channel sound card. In particular, two RME OCTAMIC XTC amplifiers and a RME MADIface USB are used. Both amplifiers digitize analog signals of up to eight channels completely synchronously. By interconnecting two RME OCTAMICs in series, up to 16 analog signals can be converted to digital values. Thus, in order to read out all signals synchronously, it is necessary to activate a delay compensation in each amplifier [27].

Similar to temporal undersampling, which causes temporal aliasing effects, spatial undersampling can lead to spatial aliasing. This effect can be observed in acoustic cameras' color maps as incorrectly detected sound sources [30]. In order to minimize spatial aliasing, there are multiple approaches possible. In [20] it is shown that an odd numbers of microphones on each ring of the array can reduce redundancy, which results in more robustness against spatial aliasing. Ring arrays in general decrease the redundancy of microphone arrays, because at a certain frequency only a few microphone pairs are affected by spatial aliasing, while others are not yet [9]. To build a sensor array, utilizing omnidirectional microphones, such as the selected condenser microphones AKG CK-92, has been shown advantageous [8, 25, 26].

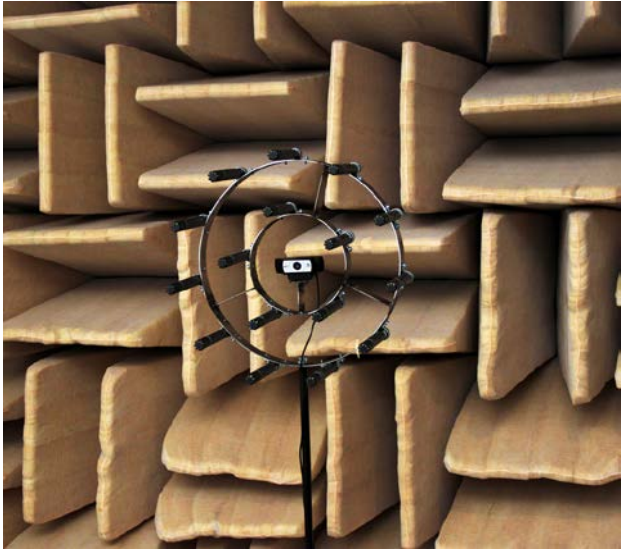

Figure 1: Developed double ring array

Using more microphones and distribute them randomly on a plane are two possible improvements of an acoustic camera's microphone array.

*B. Steered Response Power Beamforming Algorithm*

Beamforming algorithms process signals in a way, so that desired directions are enhanced, while signals from all other directions are attenuated. This chosen direction can be called steering direction, with which a defined plane can be spatially sampled. The beamformer's output, when used in this way, is known as the steered response [12]. As seen in Figure 2, the Steered Response Power Algorithm with Phase Transform weighting (SRP-PHAT) calculates a color map by summing certain values of signal pairs' weighted cross correlation (GCC). It is broadly known that a correlation results in a signal's power and thus, the steered response outputs a power. This is why the described method is known as a steered response power beamformer [11, 12].

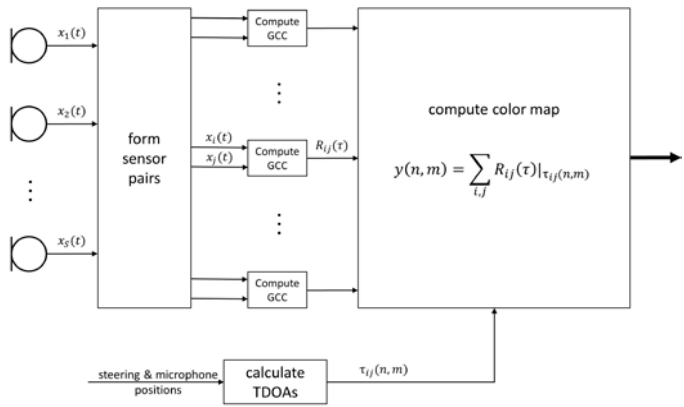

Figure 2: Schematic diagram of the SRP-PHAT algorithm, according to [17]

The mentioned GCC is similar to a regular cross correlation with the only difference, that weighted input signals are used. In order to get a PHAT weighting the Fourier transformed signals $X_1(\omega)$ and the complex conjugate of $X_2(\omega)$ are used as seen in (1). TDOA stands for time difference of arrival and it estimates the time difference of two sensors' signals. The TDOAs for a single microphone pair differ with the steering direction [11, 25].

$$\Psi_{12}(\omega) = \frac{1}{|X_1(\omega)X_2(\omega)^*|} \quad (1)$$

In [25] it has been shown, that best sound source localization results can be achieved by combining the SRP-PHAT algorithm with a constantly weighted SRP beamformer. This is because the SRP-PHAT is unable to process narrowband signals, while being very robust against reverberations and sensor self noise. The SRP method is not as robust as the SRP-PHAT algorithm, but in contrast to that, it is able to locate narrowband signals such as sine waves or spoken vowels. A combination of both is implemented by utilizing a threshold for the signal's bandwidth. In this application, the bandwidth's threshold is set to 4 kHz, which is approximately an eighth of the chosen sampling rate. A typical output image of an acoustic camera can be seen in Figure 3.

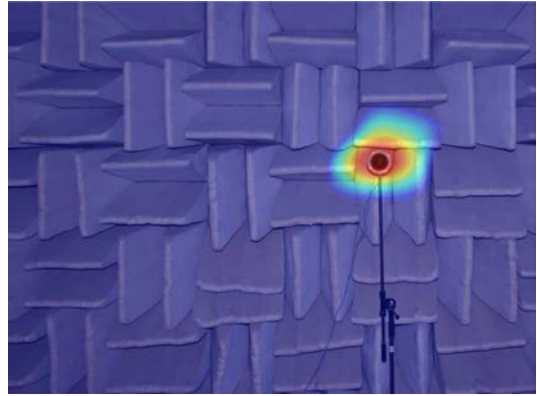

Figure 3: Resulting output image of an acoustic camera

### III. FACE DETECTION

An ideal face detection system should be able to detect all faces shown in a picture or a video frame. For this task, it should neither matter in which position or orientation the faces are, nor which age, sex or ethnical origin the people to be classified belong to. Furthermore, an ideal face detection system should be insensitive to lighting changes or other external influences [16].

In OpenCV, a face library is implemented, which provides pre-trained face detectors as well as the possibility to train own classifiers [2, 22]. Pre-trained classifiers for Haar-like and local binary pattern features support frontal face, facial landmark and whole person detection. If a self-trained classifier is used, several thousand pictures of non-faces and faces should be collected. A good training set considers faces with differences in age, sex, ethnical origin, facial hair, lighting and hairstyle [6, 13, 21]. Because of the complexity of the training process, only pre-trained classifiers for frontal faces are utilized in this contribution.

While using the later described face detection algorithms utilizing Haar-like or local binary pattern features, often multiple faces result from one facial image. If these detected faces are located in a specific area close to each other, they are averaged in size and position to merge them into one detection result. This avoids multiple detections for a single person and reduces false positive rates [2, 28].

Both, the Haar cascade and the local binary pattern classifier are implemented as cascaded classifiers to quickly reject non-faces but still keep a high accuracy for positive results (see Figure 4).

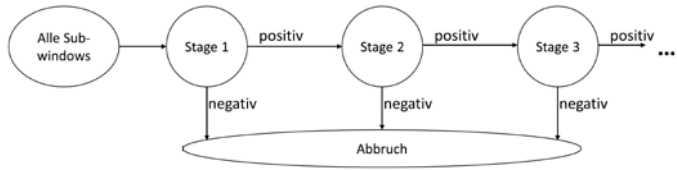

Figure 4: Schematic description of the detection cascade, according to [35]

### A. Haar Cascade Classifier

The Haar cascade classifier is a quite easy face detection method, and is therefore a very good basis for more complex algorithms. With a huge dataset, many different objects can be trained, e.g. faces, cars or whole persons. In order to classify images, Haar-like features (see Figure 5) are used and calculated extremely efficiently with integral images. Thus, regional knowledge can be considered. As very common, the face detector introduced in [35] can only handle grayscale images [6, 15, 35].

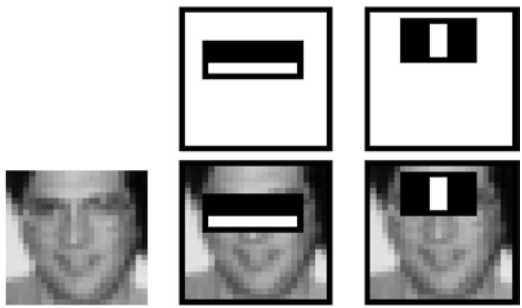

Figure 5: First two Haar-like features, according to [35]

In pictures with a resolution of 24x24 Pixel, more than 180,000 different Haar-like features can be found. Using a machine learning (ML) algorithm, the 6,061 most important features can be chosen and organized in a cascade structure. Training the chosen features $f_j$ results in a threshold $\theta_j$ and a parity $p_j$. With features being all pixel values added in black blocks and subtracted from the sum of pixel values in the white blocks, the weak classifier can be described as [35]

$$h_j(x) = \begin{cases} 1, & \text{if } p_j f_j(x) < p_j \theta_j \\ 0, & \text{else} \end{cases} \quad (2)$$

In the first five stages of the cascade, 1, 10, 25, 25 and 50 of these weak classifiers are utilized to differ between face and non-face images. The number of weak classifiers per stage is determined by a defined false positive rate, which has to be achieved in each stage. Thus, it can be imagined that in the first few stages, only few features are necessary to get to this rate, but at the very last stage, very many are needed. Stages are added as long as a total false positive rate is met [35].

Using local binary patterns, more sophisticated Haar-like features or additional non-frontal face detectors, higher detection accuracies can be achieved [13, 28, 35].

### B. Local Binary Pattern Classifier

Local binary patterns (LBPs) describe local relationships between neighboring pixels in a 3x3 environment. Starting in the top left corner and proceeding clockwise, the pixels' grayscale values are compared to the center pixels'. If the value of the center pixel at *(n,m)* is bigger than the neighbor's value, a 0 results, 1 else. These binary values can be put together and converted into a grayscale value of 0…255. Formally, this can be written as [16, 28]

$$LBP(n,m) = \sum_{n=0}^{7} s(i_n - i(n,m)) \cdot 2^n \quad (3)$$

With *s(x)* being 0, if *x<0*, 1 else. It is quite clear, that the LBP is invariant to monotone grayscale transformations, e.g. changes in brightness or contrast. LBPs contain most information for maximum two changes between 0 and 1. Examples therefore can be seen in Figure 6 [28].

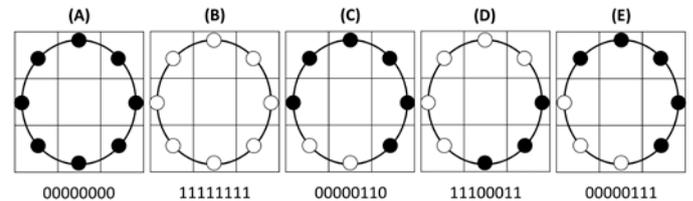

Figure 6: LBPs for points (A, B), lines (C), edges (D) and corners (E), according to [23]

Similar to the stage structure of a Haar cascade classifier, a weak classifier is formed by using a gray value histogram. With $H_n(X)$ being the classifier for the *n*-th stage and $h_{(n,m)}(x)$ being the histogram value for listing *x*, the weak classifier can be described as in (4). The LBP classifier is not only faster than the Haar cascade classifier, but theoreticaly even more precise [3, 28].

$$H_n(X) = \sum_{(n,m) \in W_n} h_{(n,m)}(x) \quad (4)$$

### IV. FACE RECOGNITION

Even though face recognition is a much more challenging task than face detection is, today's face recognition systems are, at least under optimal conditions, very reliable [2, 14]. Thus, many of today's applications use these methods to identify people in images, e.g. Facebook's Gallery or Apple iPhoto [32]. When classifying, problems mostly occur due to variations in light, perspective or facial expression [14, 32]. Fur-thermore, similar looking individuals, e.g. father and son or twins, can cause uncertainties when differing them [15].

In order to get a higher face recognition accuracy, different approaches can be imagined. In general, it is recommended to use large datasets with many variations in pose, age and lighting conditions for training the model [3, 13, 32]. Another possibility to improve the recognition performance is to use infrared lighting to avoid shades or other disruptions. Furthermore, additional features, which only occur under invisible light, e.g. freckles and pigmentation, can be used to recognize faces [15].

In OpenCV, face recognizers using principal component analysis (Eigenfaces), linear discriminant analysis (Fisherfaces) and local binary pattern histograms are implemented [23]. Utilizing any of these methods, the face recognizer has to be trained with own face-images to differentiate between individuals. The classification is done by comparing the images' features in a high-dimensional feature space with a K-nearest neighbor algorithm [6].

### A. Eigenface Classifier

The Eigenface classifier uses a principal component analysis (PCA) to reduce the dimensionality of the images. Utilizing a PCA, $E$ eigenvectors with the highest eigenvalues can be selected to describe the given dataset. These eigenvectors span a quite low-dimensional face space, in which every image can be projected. Because the PCA's eigenvectors, after reshaping them into an image format, look very much like faces (see Figure 7), they are called Eigenfaces [4, 34].

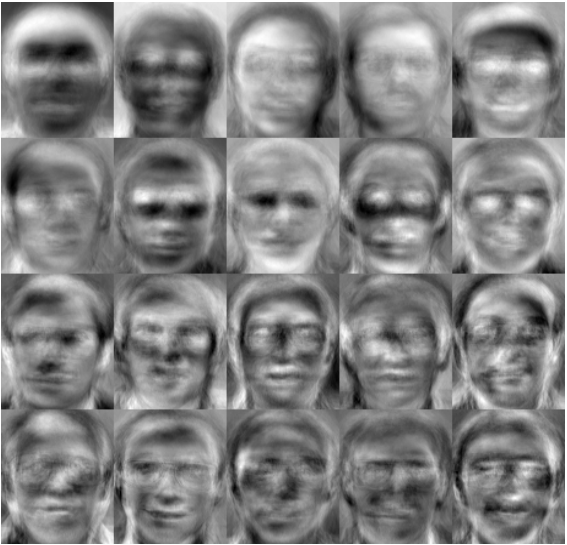

Figure 7: First 20 Eigenfaces of the AT&T face dataset, according to [23]

Using the extracted Eigenfaces, unknown faces can be reconstructed (see Figure 8). This gives a good feeling about how many principal components are necessary to distinguish between individuals. Usually a number of 40 to 80 should be sufficient, but, depending on the dataset, sometimes up to 300 Eigenfaces should be used [13, 23, 36]. Figure 8 shows that the original face can be recognized starting at 20 Eigenfaces.

In order to calculate a dataset's $(\Gamma_1, \Gamma_2, \Gamma_3, ..., \Gamma_R)$ eigenvectors efficiently, its vectorized mean image $\Psi$ and the differences $\Phi_i = \Gamma_i - \Psi$ are needed. Given a face matrix $A = [\Phi_1\ \Phi_2\ ...\ \Phi_R]$, the covariance matrix of all face-images can be calculated as [34]

$$C = \frac{1}{R}\sum_{r=1}^{R} \Phi_r \Phi_r^T = AA^T \quad (5)$$

This Matrix $C$ inherits a dimension of $N^2 \times N^2$ (for face images with a resolution of $N \times N$), which means that $N^2$ eigenvektors $u_k$ have to be determined. This requires high computational resources and thus, it is unsuitible for real-time applications. For the common case that $R<N^2$, using a workaround by building a $R \times R$ dimensional Matrix $L = A^T A$, only $R$ eigenvectors $v_l$ are to be calculated. Utilizing them, the originally disired eigenvektors $u_l$ can be determined as [19, 34]

$$u_l = \frac{1}{R}\sum_{r=1}^{R} v_{lr} \Phi_r, \quad l = 1, ..., R \quad (6)$$

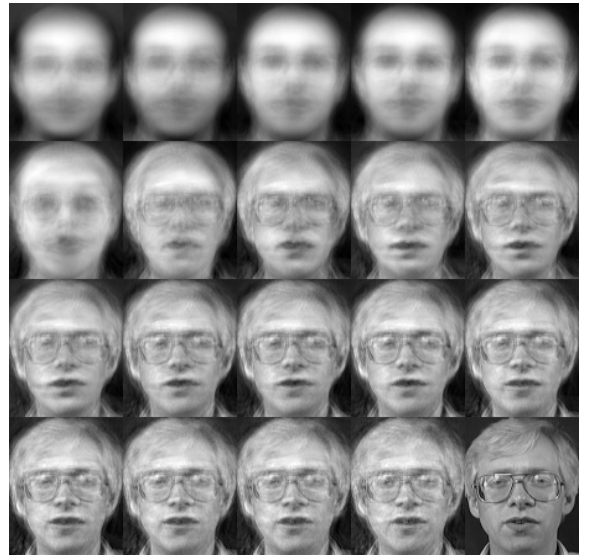

Figure 8: Reconstruction of a face using Eigenfaces, according to [23]. Number of utilized Eigenfaces (top left to bottom right): 1, 2, 3, 4, 5, 10, 20, 30, 40, 50, 60, 80, 90, 100, 125, 150, 175, 200, 300, original face

Every new facial image can be disassembled into $E$ eigenvectors as seen in (7). The resulting weights can be used to descibe a weight-vector $\Omega = (\omega_1, \omega_2, ..., \omega_E)^T$ [34].

$$\omega_k = u_k^T \cdot (\Gamma - \Psi) \quad (7)$$

Appling an Euclidian distance measure and the K-nearest neighbor method, faces can be classified [19, 34].

### B. Fisherface Classifer

Dimensionality reduction by linear discriminant analysis (LDA) can counter the PCA's disadvantage, not considering any class dependencies while projecting images in a feature space. Using Fisher's linear discriminant analysis (FLD), the classes stay linearly separable, which makes classification easier and more reliable, especially for changes in lighting condi-tions. Thus, *E* orthonormal vectors describe a matrix **W**, so that it maximizes between-class scatter (see (8)) but minimizes within-class scatter (see (9)). For both, matrices $S_B$, $S_W$ have to be defined

$$S_B = \sum_{i=1}^{C} N_i \cdot (\Psi_i - \Psi)(\Psi_i - \Psi)^T \qquad (8)$$

$$S_W = \sum_{i=1}^{C} \sum_{x_k \in X_i} (x_k - \Psi_i)(x_k - \Psi_i)^T \qquad (9)$$

where *C* is the number of different classes, $N_i$ the number of test images of class $X_i$ and $\Psi_i$ is its mean image. For face recogniton tasks, the LDA projection $W_{opt}'$ can be written as in (10). Using for example a PCA, the dimension reduces to *N–C*, while the FLD reduces it further to *C–1* [4].

$$W_{opt}^T{'} = W_{fld}^T W_{pca}^T \qquad (10)$$

with

$$W_{pca}^T = \arg \max_W |W^T S_T W| \qquad (11)$$

$$W_{fld}^T = \arg \max_W \frac{|W^T W_{pca}^T S_B W_{pca} W|}{|W^T W_{pca}^T S_W W_{pca} W|} \qquad (12)$$

The Fisherface methods provides better handling for background and lighting, than Eigenfaces do [13]. Furthermore, Fisherfaces are much more reliable when using a small training set or faces differing heavily from the training data, e.g. wearing glasses or facial expressions [4, 23].

*C. Local Binary Pattern Histogram Classifier*

The classification using local binary pattern histograms (LBPH) is quite similar to the face detection with LBP. Differences are that in order to identify individuals, a K-nearest neighbor method is utilized and the LBP operator can be extended to get results that are more reliable. For this, multiple approaches are possible. Instead of considering eight direct neighbors, *P* neighboring pixels on a radius *R* can be used for the generalized $LBP_{P,R}$ operator [1, 15, 23]. Another option is to use a multi-block LBP to compare the avarage gray scale values of neighboring pixel blocks with the avarage of a centered region [15].

The LBPH classifier's main disadvantage is that it is quite slow and therefore unsuitable for fluent video playback in real-time situations [1]. Thus, as described in the following chapter, it cannot be used in the implemented application.

V.  APPLICATION OF A FACE RECOGNITION SYSTEM

In order to compensate changes in lighting, face rotation, background and hairstyle, some preprocessing steps are taken before recognizing faces (see Figure 9). It can be assumed that this allows to apply the classifiers not only to constrained environments, but also to any [34].

As a suitable face detection algorithm, the Haar cascade classifier is chosen. Even though, OpenCV's LBP classifier shows in practice an approximately 61 % faster processing time, it is less accurate and is unable to detect faces reliably in an artificially lighted office room. In comparison to the LBP classifier, the Haar cascade classifier is a little slower, but still capable of consistently classify pictures into faces and non-faces. In order to detect the left and right eye in an image, the corresponding Haar cascade eye detectors are utilized.

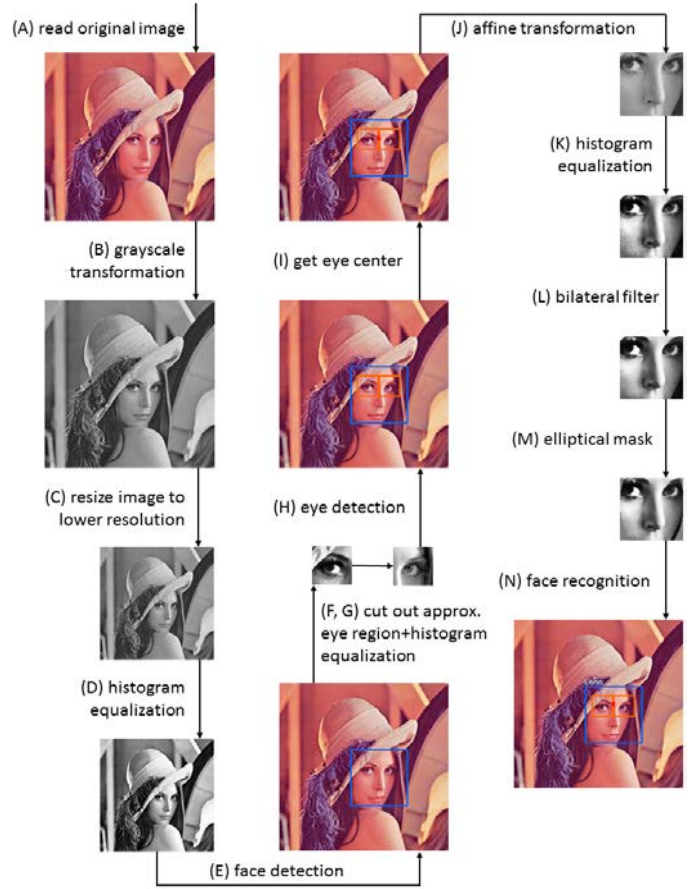

Figure 9: Preprocessing steps for face recognition, according to [3, 23]

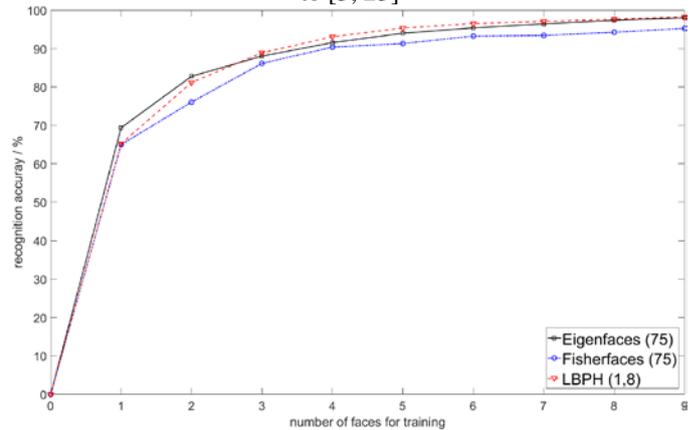

Figure 10: Recognition accuracy vs. number of training faces

To find the best face recognizer for the described application, several tests are considered. In particular, these tested the accuracy, training duration, model size and recognition speed over number of components (Eigenfaces/Fisherfaces) and classes in the training set as well as their total recognition accuracy over the number of training images (see Figure 10).

It can be seen, that, after applying the preprocessing steps, the Eigenface method constantly outperforms the Fisherface method, even though. For three and more faces, the LBPH is slightly more accurate than the Eigenface method. The test also shows that the training duration of Eigenfaces is for relatively small numbers of components slightly lower than the Fisherface method, while this changes for larger numbers of components. The training duration of LBPH is by far longer, especially when using larger radii and more neighbors. This also shows in comparing them at model sizes. LBPHs' model sizes are a lot larger than PCAs' and LDAs' models, which show the same magnitude, even though the LDA models are smaller. The Fisherfaces' recognition speeds are, starting at equally many classes and components, slightly higher than the Eigenfaces'. Before that point, both are almost identically high and about five times bigger than the LBPH's recognition speed. As seen in Figure 11, the Fisherfaces' recognition accuracy reaches its maximum approximately at the number of components being equal to the number of classes. This can be explained with the amount of class in the dataset. As described in section IV.B, the number of components is limited to $C-1$, which means adding additional components would not increase the recognition accuracy. Similar to this, the Eigenfaces' recognition accuracy reaches its maximum at $R$, the number of images in the training set (see section IV.A). Using five classes times ten images (minus two for testing), this limit is reached at approximately 40 components.

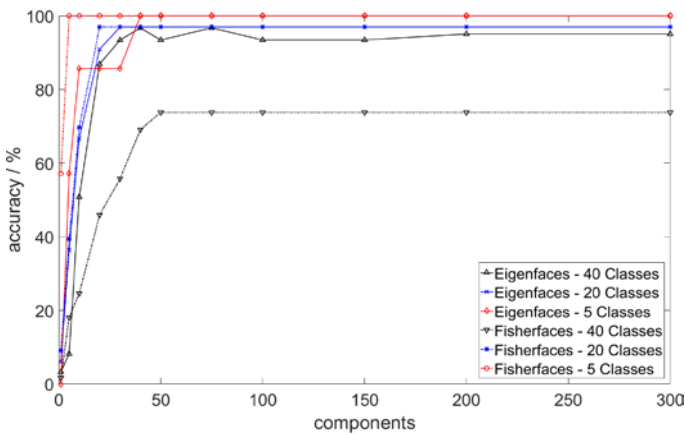

Figure 11: Recognition accuracy vs. number of components

These performance measures suggest using an Eigenface classifier because of its higher recognition accuracy and else similar properties. Further tests has shown, that using approximately 30 components, show best recognition results. This matches literatures suggestions [13, 23, 36]. [4] recommends not using the first three Eigenfaces to achieve even higher accuracies. Unfortunately this option is not supported by OpenCV [24].

In the final application, images for a face dataset have been collected. There, four individuals with approximately 1400 images are considered. The system implemented in OpenCV runs in real-time, providing approximately 15-18 frames per second.

## VI. FUSION OF THE LOCALIZATION RESULTS FOR SPEAKER IDENTIFICATION

A fusion of the localization results of sound source localization and face recognition is able to enhance the reliability of a speaker detection system and enables it to identify the speaker. This can be used in smart rooms, improved speaker tracking for videoconferencing or applications for ambient assisted living [7, 8, 29]. Furthermore, an extension towards gesture recognition for human machine interaction is possible [7, 10]. Therefore, a speaker identification algorithm is developed and introduced in this contribution.

In order to track and identify speakers, reliable sound source localization and face recognition are necessary. The sound source and therefore the potential speaker, is located by finding the color map's maximum. The face recognition provides a specific localization as well, but sometimes there are some uncertainties, which have to be eliminated. These could be falsely positive detected and recognized faces or wrongly classified individuals. To overcome these problems, three recognized faces are compared. If all of them are classified as the same person, the result is shown at the face's new position. Another possibility is that the recognized face matches one of the two previously identified individuals and is in approximately the same localization area as the currently detected. This results in a certain recognition at the currently classified face's position, too. If both options do not apply, no recognition result is being displayed and it is ignored like there never has been a face in the image. Using both, the speaker and face localization results, the overall localization and identification can be achieved as seen in Figure 12. The estimated outcome can differentiate between following: no result, identified speaker, face only, unknown speaker or loudspeaker, loudspeaker and known face at two different positions. Whenever possible, the localization position is chosen to the face's location, because it is well known that optical tracking algorithms have better spatial resolution than acoustic localization techniques [8].

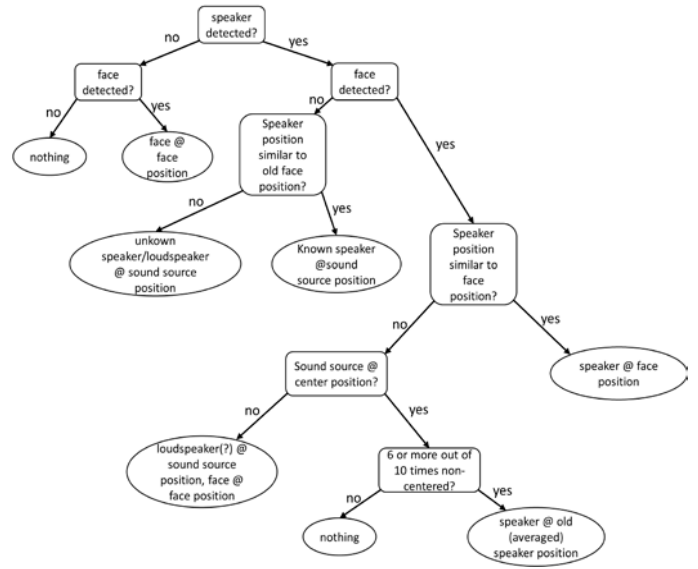

Figure 12: Decision tree for localization result

Figure 13 shows a possible output image of the implemented speaker identification system. There, a speaker's face and sound source location are detected and merged for a more precise localization and identification of the sound source. The red circle marks the speaker's approximate position.

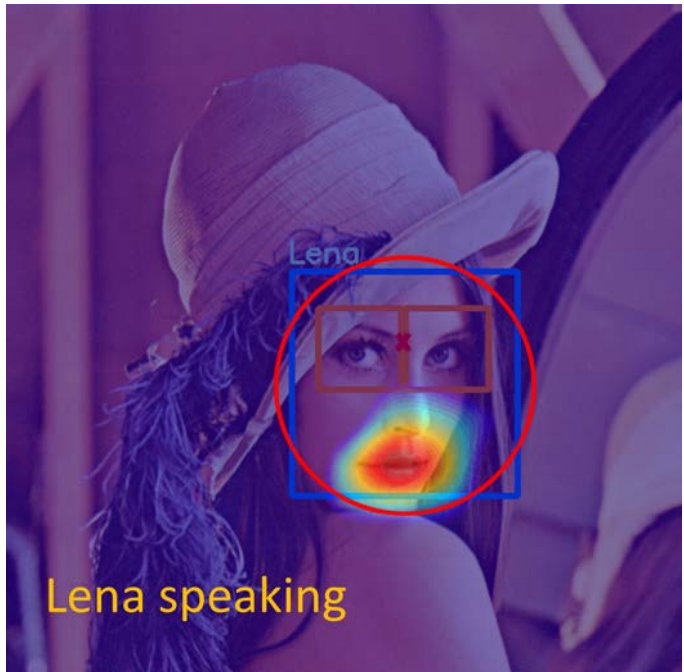

Figure 13: Result of the speaker identification system

## VII. Conclusion

This contribution briefly explains the basics of an acoustic camera and shows, why it makes sense to use a double ring array with an odd number of microphones. Additionally, it gives an overview of the implemented sound source localization methods. It can be shown, that a combination of SRP and SRP-PHAT algorithms is desirable for speech localization. Furthermore, this contribution gives an introduction to face detection and recognition methods. It is shown that Haar cascade classifiers outperform local binary pattern classifiers, when detecting faces. Similar, it is pointed out that for a face recognition system, Eigenfaces should be preferred to Fisherfaces and local binary pattern histograms. Finally, an algorithm for the fusion of localization results is introduced. This combines sound sources localization and face detection to identify speakers reliably.

## References

[1] Ahonen, T., Hadid, A. u. Pietikäinen, M.: Face Recognition with Local Binary Patterns. In: European Conference on Computer Vision 2004, Springer-Verlag Berlin, pp. 469–481
[2] Arubas, E.: Face Detection and Recognition (Theory and Practice), 2013. http://eyalarubas.com/face-detection-and-recognition.html, accessed on: 16.02.2017
[3] Baggio, D. L.: Mastering OpenCV with practical computer vision projects. Step-by-step tutorials to solve common real-world computer vision problems for desktop or mobile, from augmented reality and number plate recognition to face recognition and 3D head tracking. Birmingham: Packt Publ 2012
[4] Belhumeur, P. N., Hespanha, J. P. u. Kriegman, D. J.: Eigenfaces vs. Fisherfaces: Recognition Using Class Specific Linear Projection. In: IEEE TRANSACTIONS ON PATTERN ANALYSIS AND MACHINE INTELLIGENCE, VOL. 19, NO. 7, JULY 1997
[5] Bergh, T. F., Hafizovic, I. u. Holm, S.: Multi-speaker voice activity detection using a camera-assisted microphone array. IWSSIP Bratislava 2016. The 23rd International Conference on Systems, Signals and Image Processing : Bratislava, Slovakia, 23-25 May 2016 : proceedings. Piscataway, NJ: IEEE 2016
[6] Bradski, G. R. u. Kaehler, A.: Learning OpenCV. Software that sees. 1005 Gravenstein Highway North, Sebastopol, CA 95472.: O'Reilly 2008
[7] Busso, C., Georgiou, P. G. u. Narayanan, S.: REAL-TIME MONITORING OF PARTICIPANTS' INTERACTION IN A MEETING USING AUDIO-VISUAL SENSORS. In: Acoustics, Speech and Signal Processing, 2007. ICASSP 2007. IEEE International Conference, pp. 685–688
[8] Busso, C., Hernanz, S., Chu, C.-W., Kwon, S.-i., Lee, S. U., Georgiou, P. G., Cohen, I. u. Narayanan, S.: Smart Room: Participant and Speaker Localization and Identification. Proceedings / 2005 IEEE International Conference on Acoustics, Speech, and Signal Processing. May 18 - 23, 2005, Pennsylvania Convention Center/Marriott Hotel, Philadelphia, Pennsylvania, USA. Piscataway, NJ: IEEE Operations Center 2005, pp. 1117–1120
[9] Clénet, B.: Circular Microphone Array Based Beamforming And Source Localization On Reconfigurable Hardware, Graz University of Technology. Master Thesis. Graz, Austria 2010
[10] Dai, J., Wu, J., Saghafi, B., Konrad, J. u. Ishwar, P.: Towards privacy-preserving activity recognition using extremely low temporal and spatial resolution cameras. 2015 IEEE Conference on Computer Vision and Pattern Recognition workshops (CVPRW). 7 - 12 June 2015, Boston, MA. Piscataway, NJ: IEEE 2015, pp. 68–76
[11] DiBiase, J., Silverman, H. u. Brandstein, M.: Robust Localization in Reverberant Rooms. In: Brandstein, M. u. Ward, D. (Hrsg.): Microphone arrays. Signal processing, techniques and applications. Engineering online library. Berlin: Springer 2001, pp. 157–180
[12] DiBiase, J. H.: A High-Accuracy, Low-Latency Technique for Talker Localization in Reverberant Environments Using Microphone Arrays, Brown University PhD Thesis. Providence, Rhode Island 2000
[13] Howse, J., Puttemans, S., Hua, Q. u. Sinha Utkarsh: OpenCV 3 Blueprints. Community experience distilled. s.l.: Packt Publishing 2015
[14] Huang, T., Xiong, Z. u. Zhang, Z.: Face Recognition Applications. In: Li, S. Z. u. Jain, A. K. (Hrsg.): Handbook of Face Recognition. London: Springer-Verlag London Limited 2011, pp. 617–638
[15] Li, S. Z. u. Jain, A. K. (Hrsg.): Handbook of Face Recognition. London: Springer-Verlag London Limited 2011
[16] Li, S. Z. u. Wu, J.: Face Detection. In: Li, S. Z. u. Jain, A. K. (Hrsg.): Handbook of Face Recognition. London: Springer-Verlag London Limited 2011, pp. 277–303
[17] Lombard, A. J. V.: Localization of Multiple Independent Sound Sources in Adverse Environments. Lokalisierung mehrerer unabhängiger Schallquellen in widrigen Umgebungen, Universität Erlangen-Nürnberg Promotionsschrift. Erlangen-Nürnberg 2012
[18] Mabande, E.: Robust Time-Invariant Broadband Beamforming as a Convex Optimization Problem. Robuste zeitinvariante Breitband-Keulenformung als konvexes Optimierungsproblem, Friedrich-Alexander-Universität Erlangen-Nürnberg. Erlangen-Nürnberg 2014
[19] Martinovský, F. u. Wagner, P.: Gesichtserkennung mit Eigenfaces. http://www.bytefish.de/pdf/eigenfaces.pdf, accessed on: 16.02.2016
[20] Möser, M. (Hrsg.): Messtechnik der Akustik. Berlin: Springer 2010
[21] OpenCV: Cascade Classifier Training — OpenCV 2.4.13.2 documentation. http://docs.opencv.org/2.4.13.2/doc/user_guide/ug_traincascade.html, accessed on: 31.03.2017
[22] OpenCV: OpenCV: Face Detection using Haar Cascades. http://docs.opencv.org/3.1.0/d7/d8b/tutorial_py_face_detection.html, accessed on: 16.02.2016
[23] OpenCV: OpenCV: Face Recognition with OpenCV. http://docs.opencv.org/3.1.0/da/d60/tutorial_face_main.html#tutorial_face_lbph, accessed on: 16.02.2016
[24] OpenCV Community, 2014. http://answers.opencv.org/question/26188/eigenface-algorith-can-be-improved/, accessed on: 01.03.2017
[25] Reschke, J.: Implementation of a Steered Response Power Acoustic Camera, Ostbayerische Technische Hochschule Regensburg project report. Regensburg 2016
[26] Reschke, J.: Aufbau und Test eines mehrkanaligen Audioaufnahmesystems für eine akustische Kamera, Ostbayerische Technische Hochschule Regensburg project report. Regensburg 2016
[27] RME: Bedienungsanleitung OctaMic XTC. The Professional's Multiformat Solution. http://www.rme-audio.de/download/octamicxtc_d.pdf, accessed on: 08.06.2016
[28] Rodriguez, Y.: Face Detection and Verification using Local Binary Patterns, École polytechnique fédérale de Lausanne Promotionsschrift. Lausanne, Schweiz 2006
[29] Rozgic, V., Busso, C., Georgiou, P. G. u. Narayanan, S.: Multimodal Meeting Monitoring: Improvements on Speaker Tracking and Segmentation through a Modified Mixture Particle Filter. In: Multimedia Signal Processing, 2007. MMSP 2007. IEEE, pp. 60–65
[30] Scholte, R., Roozen, B. u. Lopez, I.: On Spatial Sampling and Aliasing in Acoustic Imaging. In: Twelfth International Congress on Sound and Vibration. 2005
[31] Sigl, J. u. Scheucher, R.: Acoustic Imaging of Sound Sources with a student-designed Acoustic Camera. AMERICAN JOURNAL OF UNDERGRADUATE RESEARCH 6 (2007) 2
[32] Stieler, W.: Die Schutzbrille. Technology Review 02 (2017), pp. 82–83
[33] Szeliski, R.: Computer vision. Algorithms and applications. Texts in computer science. London: Springer 2011
[34] Turk, M. u. Pentland, A.: Eigenfaces for Recognition12. Journal of Cognitive Neuroscience 3 (1991) 1
[35] Viola, P. u. Jones, M.: Rapid Object Detection using a Boosted Cascade of Simple Features. In: ACCEPTED CONFERENCE ON COMPUTER VISION AND PATTERN RECOGNITION 2001
[36] Willow Garage: OpenCV Face Module. OpenCV 2015